\documentclass{article}

\usepackage[utf8]{inputenc} 
\usepackage[T1]{fontenc}    
\usepackage{hyperref}       
\usepackage{url}            
\usepackage{booktabs}       
\usepackage{amsfonts}       
\usepackage{nicefrac}       
\usepackage{microtype}      
\usepackage{graphicx}
\usepackage{doi}
\usepackage[disable]{todonotes}
\usepackage{snaptodo}
\usepackage{subcaption}
\usepackage{comment}
\usepackage{amsmath}
\usepackage{bm}
\usepackage{arxiv}

\newcommand{\alvin}[1]{\todo[color=green!40]{#1}}

\title{Examining Pathological Bias in a Generative Adversarial Network Discriminator: A Case Study on a StyleGAN3 Model}

\date{} 					

\author{ \href{https://orcid.org/0000-0002-6503-2703}{\includegraphics[scale=0.06]{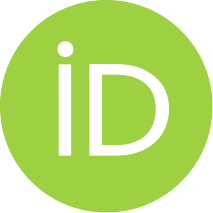}\hspace{1mm}Alvin {Grissom~II}} \\
	Department of Computer Science\\
	Haverford College, USA\\
	\url{agrissom@haverford.edu} \\ 
	\And
	{Ryan F.~Lei} \\
	Department of Psychology\\
	Haverford College, USA\\
	\url{rlei1@haverford.edu} \\ \\
 	\And
  {Matt Gusdorff} \\
	Department of Computer Science\\
	Haverford College, USA\\
	 \\
 \And
	{Jeová Farias Sales Rocha Neto} \\
	Department of Computer Science\\
	Bowdoin College, USA\\
	\url{j.farias@bowdoin.edu} \\ 
 \And
 {Bailey Lin} \\
	Department of Computer Science\\
	Haverford College, USA\\ 
  \And
 {Ryan Trotter} \\
	Department of Computer Science\\
	Haverford College, USA\\
}

\renewcommand{\shorttitle}{Examining Pathological Bias in a GAN Discriminator}

\hypersetup{
pdftitle={Examining Pathological Bias in a Generative Adversarial Network Discriminator: A Case Study on a StyleGAN3 Model},
pdfauthor={Alvin Grissom II, et al.},
pdfkeywords={computer vision, generative adversarial networks, bias, gans, deepfakes},
}

\begin{document}
\maketitle
\begin{abstract}
    Generative adversarial networks generate photorealistic faces that are often indistinguishable by humans from real faces.  As part of their training procedure, GANs train both a generator and discriminator network.  We find that the discriminator in the pre-trained StyleGAN3 model, a popular GAN network, systematically stratifies scores by both image- and face-level qualities and that this disproportionately affects images across gender, race, and other categories.  We examine the discriminator's bias for color and luminance across axes perceived race and gender; we then examine axes common in research on stereotyping in social psychology.
\end{abstract}
    


\section{Introduction}
Generative adversarial networks (GANs) have seen widespread adoption in machine learning, especially in computer vision applications.  These ``generative'' models are capable of producing artificial images in many instances indistinguishable from the real thing.  The most common use of these networks is that of artificial face generation.  These so-called ``deepfakes'' have been used in a number of research and commercial applications. 

With their proliferation, however, have come predictable problems of bias in their generation.  All such models are trained on large datasets.  Several pre-trained models for StyleGANs 2 and 3 are trained on the Flickr (FFHQ) dataset.\footnote{\url{https://github.com/NVlabs/ffhq-dataset}}  These data are known to have imbalances in racial, ethnic, and gender representation, and this likely leads to bias in the kinds of faces they generate, but while it is often assumed that learned models merely reflect these biases, this is by no means obvious (and indeed we find evidence that this is not the case). Latent model biases in complex networks can be difficult to detect.

Our work examines one crucial part of this puzzle.  We investigate StyleGAN3-r's discriminator, which is used during training as the generator's adversary.  We want to know how the discriminator discriminates.

We use an official pre-trained StyleGAN3 model provided by NVIDIA\footnote{\url{https://github.com/NVlabs/stylegan3}} and examine how the trained discriminator classifies faces by examining salient properties of the images.  In addition to providing a general example of how researchers can examine biases inside of GANs despite their architectural complexity, we have three main findings, with several nuances: (1) we find stark correlations with abstract image properties of color and luminance which, counterintuitively, \textit{do not} merely reflect color distributions in the training data; (2) we find that the the discriminator is systematically biased against certain racialized groups, especially Black men; and (3) we find that men  with long hair receive systematically lower ``realness'' scores than men short hair.  In short, the GAN discriminators are ``racist'' in the sense they systematically under-score the faces of certain races.\footnote{We adamantly do not reify biological notions of race.  For the purposes of this study, ``race'' may be read as ``perceived race'' from three categories: Black, Asian, and white.  We do not claim that these are biologically ``real'' categories. See Section~\ref{sec:caveat} for a longer discussion on race.} 

This paper proceeds as follows: Section~\ref{sec:background} reviews background on GANs, bias in deep learning systems, and relevant psychological research; Section~\ref{sec:study1}\footnote{All human studies and surveys received IRB approval.} describes our first study on the discriminator's notion of a face in the FFHQ data; Section~\ref{sec:study2} digs deeper by examining novel faces with crowdsourced labels; and in Section~\ref{sec:modeling}, we use regression analysis to get a sense of overall trends on labeled faces.

\section{Background and Related Work}
\label{sec:background}
In this section, we review the mechanics of GANs and the StyleGAN~\cite{karras2019style} architecture that we use for our experiments and describe their relevance to the bias we study.

\subsection{Generative Adversarial Networks}
In generative modeling, given a dataset $\mathbf{X}$ with examples $\mathbf{x}\in \mathbf{X}$, we  approximate the distribution that gave rise to $\mathbf{x}$, $p_{\text{data}}(\mathbf{x})$ with a learned distribution $p_{\text{model}}(\mathbf{x})$.  If successful, $p_{\text{model}}(\mathbf{x})$ can then be used to generate novel samples---in this case, human faces. Proposed in 2014 by Goodfellow et al.~\cite{goodfellow2020generative}, generative adversarial networks approach this problem by using two artificial neural networks~\cite{lecun2015deep} that compete in a game to achieve the Nash equilibrium.

The \textbf{generator} network $G(\mathbf{z},\boldsymbol\theta^G)$ is used to draw samples from $p_{\text{model}}$. It takes a random input $\mathbf{z}$ (latent vector) sampled from a noise distribution $p(\mathbf{z})$ and generates an image-sized output. The \textbf{discriminator} network $D(\mathbf{x}, \boldsymbol\theta^D)$ network takes some sample $\mathbf{x}$ and determines whether it is real according to the examples in $\mathbf{X}$, assigning it a score. During training, the sample $\mathbf{x}$ may be drawn from the training data or generated by the generator $G$. 

During training, $G$ improves its generation ability by using the discriminator output on its generated data. The goal is to create fake data that tricks the discriminator into classifying it as real. In turn, the discriminator also learns to better distinguish between fake and real imagery by considering the data from $\mathbf{X}$ and samples from the current state of the generator. In its initial formulation, this process was stated via the following optimization problem \cite{goodfellow2020generative},
\begin{align}
\nonumber\min_{\theta^G}\max_{\theta^D}~~&\mathbb{E}_{\mathbf{x} \sim p_{\text{data}}(\mathbf{x})}[\log(D(\mathbf{x}, \theta^D))]   
\\&+ \mathbb{E}_{\mathbf{z} \sim p(\mathbf{z})}[\log(1 -  D(G(\mathbf{z}, \theta^G), \theta^D))].
\end{align}

$G$ and $D$ play this minimax game until the faces generated by $G$ are photorealistic. This framework gained unprecedented success in image generation, which led to further research improving and generalizing it. Later work led to the development of architectures that generate high resolution images \cite{karras2017progressive, brock2018large}, 3D models~\cite{wu2016learning} and text~\cite{li2019storygan}, finding applications in medicine~\cite{baek2020weakly, teramoto2020deep}, image processing~\cite{go2020deep, zhang2020image}, face recognition~\cite{zhao20183d}, physics~\cite{de2017learning} and traffic control~\cite{beery2020synthetic}.


\subsection{Face Generation and StyleGAN}
We focus our attention on generating realistic human faces, which has been challenging even for GANs due to problems such as mode collapse and the size of the networks involved. One step toward solving this issue was proposed in \cite{karras2017progressive}, where the authors propose a generator architecture that, starting from a low resolution image, progressively upsamples it and add more details to it during training. The discriminator does the reverse: it takes an RGB representation of the image and converts it into a feature map.  The discriminator then downsamples the feature map representation of the image, halving it at each step while simultaneously doubling the number of features.  Before classifying the image, the discriminators also scale the output activations by the $L_2$ norm in order to normalize the weights such that their standard deviations sum to unity. This GAN approach, called Progressive GAN (ProGAN), attained high-quality face generation results of $1024 \times 1024$ pixel size using data from the CelebA  dataset~\cite{liu2015deep}.

\alvin{Prev. paragraph can be cut if we're low on space.}
More recently, Karras et al.~\cite{karras2019style} proposed the StyleGAN network. In this  work, ProGAN's generator architecture was changed in several ways: first, the latent vector $\mathbf{z}$ is mapped onto an intermediate latent vector space $\mathbf{w}$ via a series of fully connected layers. Then, $\mathbf{w}$ is given as an input to teach progressive layers individually via an affine transform, followed by an adaptive instance normalization (AdaIN) layer. Gaussian noise is also added between each convolutional layer and non-linearity in the progressive architecture. After being trained in CelebA and  Flickr-Faces-HQ (FFHQ) datasets, Karras et al.~\cite{karras2019style} demonstrate that this network is capable of not only generating photorealistic facial images, but affords control of their style, i.e., some of its visual features such as hair length, skin color and age. 

StyleGAN2~\cite{karras2020analyzing} improves upon its predecessor by modifying its original AdaIN module, resulting in a drastic reduction of droplet-shaped artifacts present in the data generated by StyleGAN. Furthermore, instead of using the progressive methodology inherited from ProGAN, StyleGAN2 utilizes skip connections and residual modules to generate high-quality images. This change was proposed  to inhibit StyleGAN the generation of artifacts.

StyleGAN3~\cite{karras2021alias} modified StyleGAN2 by addressing the lack of translational and rotational equivariance in convolutional neural networks, reducing the reliance on absolute pixel coordinates and facilitating animated rotations.  This is accomplished by removing references to absolute pixel location from the image representation.  The discriminator of StyleGAN3e remains unchanged from that of StyleGAN2.



\subsection{Bias in Models}
There is a rich and growing literature on bias and pathologies in  deep learning models, including computer vision, which has comparatively fewer than, for example, natural language processing.  Buolamwini and Gebru~\cite{buolamwini2018gender} make a strong case for widespread gender and skin color bias in vision classification systems.  Maluleke et al.~\cite{maluleke2022studying} suggest that the biases in generated images reflect the training data, but also that the oft-used truncation trick exacerbates this. Other work has cautioned about the use of GANs for facial augmentation due to biased associations based on, e.g., skin color~\cite{jain2022imperfect, yang2022enhancing}. Furthermore, we know that deep neural networks are vulnerable to sometimes imperceptible  adversarial examples~\cite{goodfellow2014explaining} and that defy human intuition~\cite{feng-etal-2018-pathologies, shi-etal-2022-rare, kruengkrai2023revisiting}.

\subsection{A Caveat on Social Categorization}
\label{sec:caveat}
\alvin{This sec is good; can go in an epic footnote if we're low on space.}
We examine whether and how GANs may be biased, especially along gender and racial axes. However, social categorization is a complex and multifaceted psychological process that is socio-historically situated and constructed~\cite{richeson2016toward, lei2023sociohistorical}. That is, what it means to belong to a racialized group has changed over history, often to maintain a specific racial hierarchy \cite{lei2023sociohistorical}. 

Yet it is also true that though these categories are socially constructed, they have very real consequences for those who are socially subordinated. For example, those racialized as Black often experience disproportionately negative outcomes, including health \cite{pascoe2009perceived} and criminal justice \cite{ekstrom2022racial}.  Thus, when we refer to racialized groups (e.g., Black, Asian), we refer to a social psychological phenomena imposed on these groups by perceivers.

\subsection{Analogues in Psychological Research}
Our work is situated within both computer vision and psychology research.\footnote{This is not to say that we are anthropomorphizing GANs; far from it.} Analogous to our work in psychology is research asking how people come to learn what constitutes a social category and how faces are markers of these deeper conceptual representations. Specifically, research asking who people think of as the most prototypical---most representative---of a given social category provides a theoretical framework for considering salient features against which a GAN might learn biases (either as a first- or second-order consequence). Societies tend to have cultural ideologies that center men and members of dominant racial-ethnic groups in representations of social categories \cite{purdie2008intersectional}. Consequently, when people think of who is most representative of a generic person, these biases interact to center dominant groups (e.g., white men in the US) as the prototypical person. Cultural biases are learned and thus infuse people's social prototypes over the period of early childhood 
\cite{lei2021race}. 

These psychological studies provide relevant insights with respect to how GANs may be learning similar biases and deploying them when generating realistic fake images. In the same way that the inputs (e.g., the racial diversity in their social environment~\cite{hwang2021neighborhood}) children get can inform their representations of social categories, GANs are similarly susceptible to their own inputs--namely, the training data. If most faces that a GAN is trained on are from one gender or one racialized group, then the resulting face images that are generated may also reflect those biases. Indeed, the modal face that StyleGAN2 generates is that of a white woman~\cite{salminen2020analyzing, merler2019diversity}. 

Of course, the discriminator plays a critical role in deciding whether or not the generated image approximates a human face or not. This more closely parallels the psychological process of categorization. Because most infant caregivers are women, there is an early preference for female faces in infancy~\cite{quinn2002representation}, and often same-race female faces specifically~\cite{quinn2008infant}. Interestingly, the preference for female shifts to one that centers men (and specifically white men) by the early childhood period~\cite{lei2020development,lei2021race}. 

These findings from psychology generate a few possibilities for how the discriminator might discriminate on the basis of race and gender. Given that the StyleGAN's most frequently generated image is that of a white woman~\cite{salminen2020analyzing}, one possibility is that the discriminator would generally assign lower scores to images of people with darker skin tones, which is more common among Black people. A second non-mutually exclusive possibility is that the discriminator would assign lower scores to images of men with darker skin tones specifically, given that the training set of face images are primarily that of white women \cite{merler2019diversity}. 

\subsection{Overview of Current Work}

We now describe our experiments for examining the bias in the pre-trained StyleGAN 3 model.  All of our experiments investigate one question: what does the model consider a human face?  In particular, we want to know whether the model's notion of a human face---whatever it may be---is systematically biased, and if so, how.  To accomplish this, we focus specifically on the GAN's discriminator.  During the training process, the generator and discriminator are interdependent: the generator learns to generate what will fool the discriminator; thus, if the discriminator is biased, the generator will be, as well.  

In Study 1, we examine how the discriminator evaluates faces from the FFHQ dataset--the face image dataset it was trained on. In Study 2, we examine how the discriminator evaluates a novel set of faces scraped from Google.

\section{Study 1: Examining the Large Training Data}
\label{sec:study1}
To examine to what extent the discriminator is biased, we first examine whether it systematically scores faces from particular social groups (e.g., racially minoritized groups) lower scores vs. others.

\subsection{Discriminator Bias for Lighter Skin Tones}

\begin{figure*}[t!]
\centering
\begin{subfigure}[]{0.33\linewidth}
 \includegraphics[width=\textwidth]{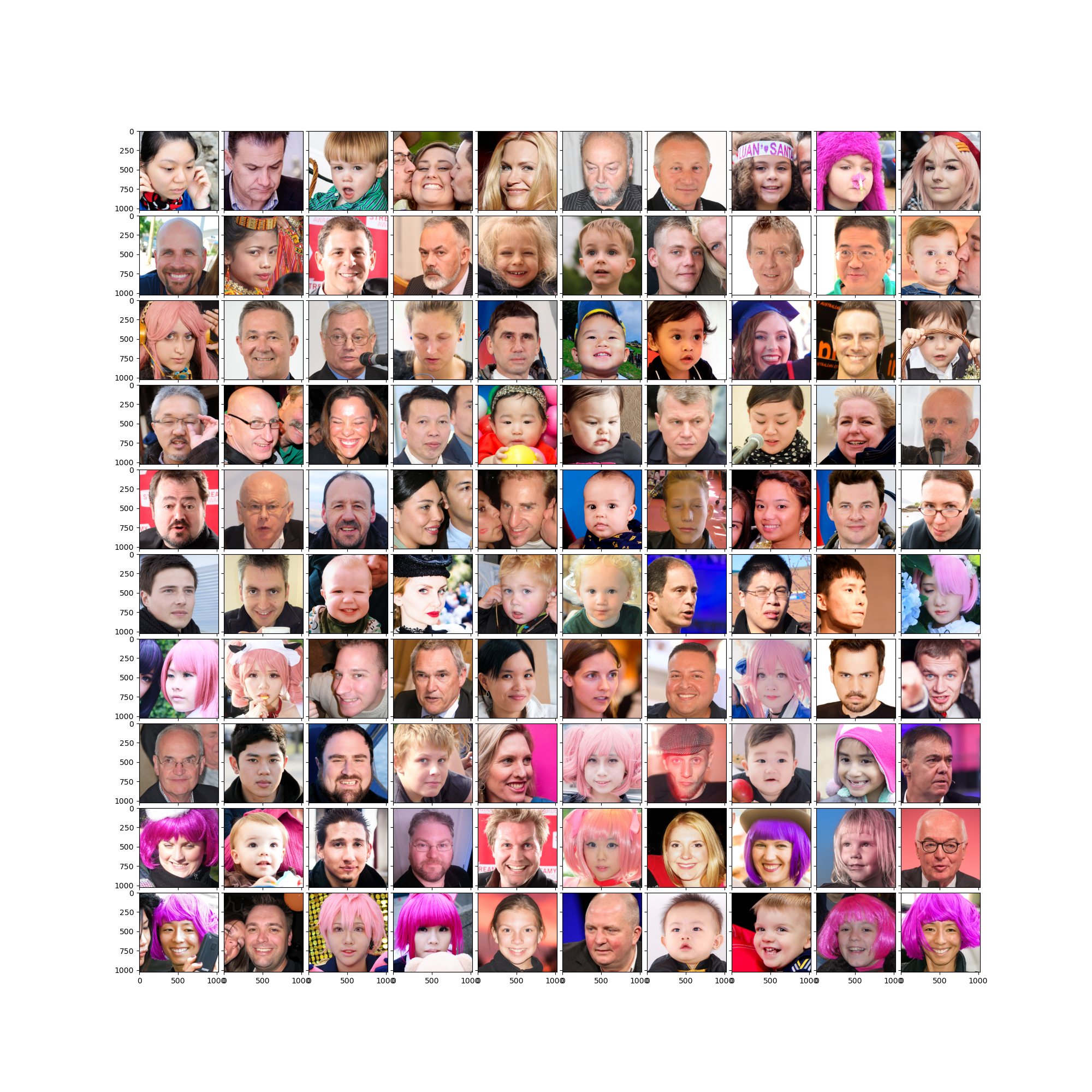}
\caption{100 highest-scoring faces}
 \label{fig:top100}
\end{subfigure}%
   \hfill
\begin{subfigure}[]{0.33\linewidth}
\includegraphics[width=\textwidth]{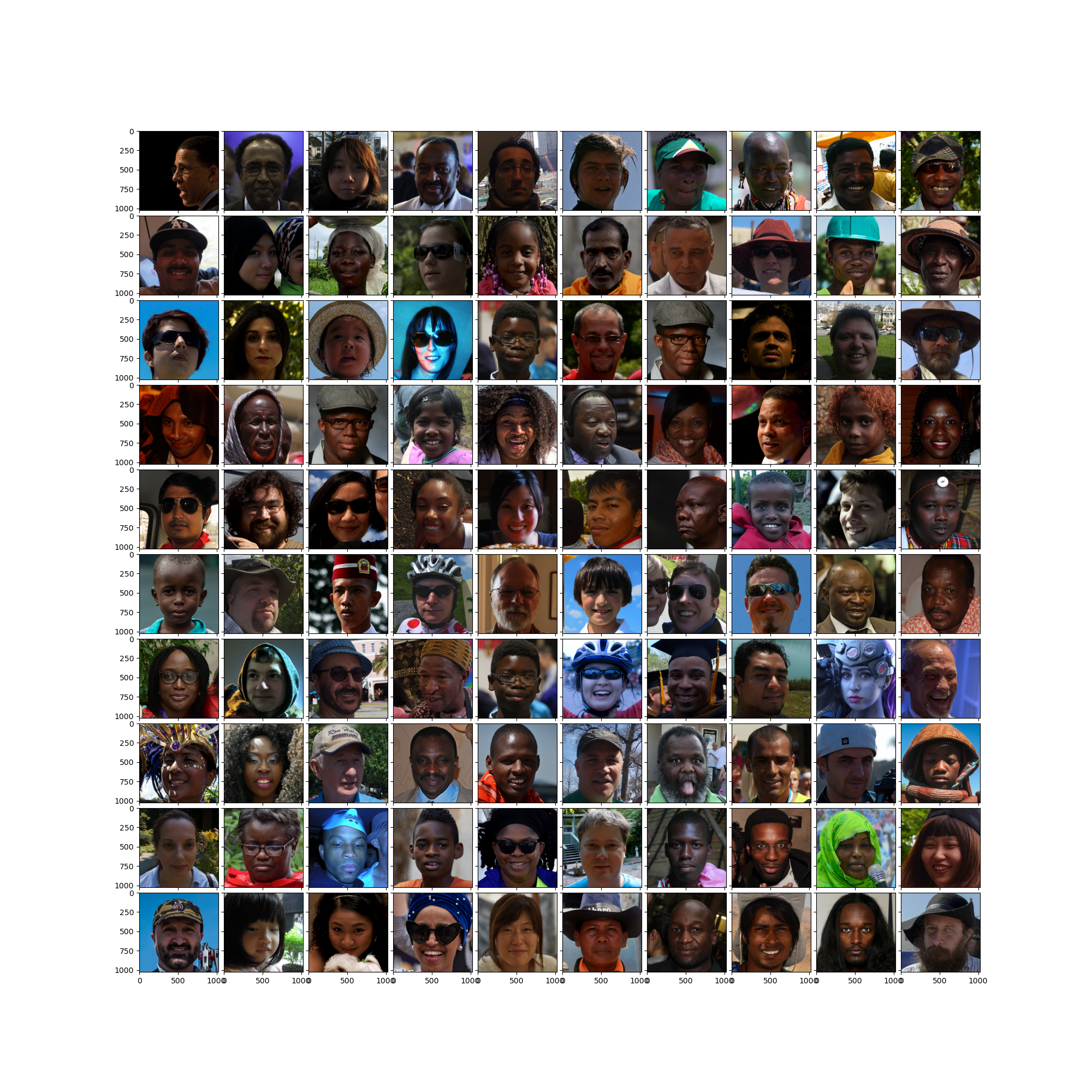}
\caption{100 lowest-scoring faces}
    \label{fig:least100}
\end{subfigure} 
\begin{subfigure}[]{0.33\linewidth}
\centering
\includegraphics[width=\textwidth]{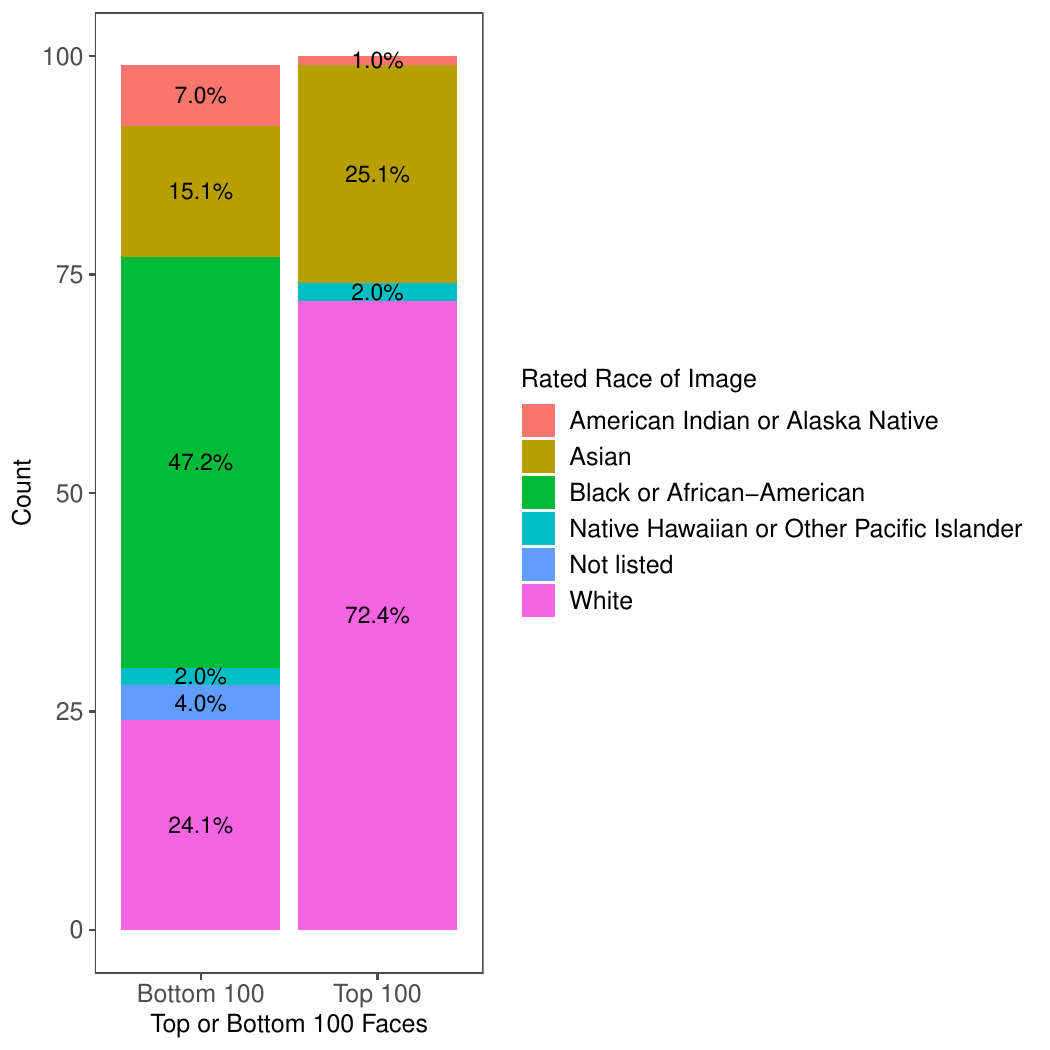}
\caption{Crowdsouced racial categorizations}

\label{fig:racial categorization}
\end{subfigure}

\caption{While there are some examples of faces with lighter skin receiving a low score in a dark image, the stark bias for lighter-skinned faces and against darker-skinned ones is clear. There is n propensity to favor the color pink. None of the top 100 are labeled Black.
}
\label{fig:topbottom100}
\end{figure*}

 To test  discriminator bias on a large dataset, we score every image in the FFHQ dataset and sort by discriminator score, examining the top-scoring and lowest-scoring 100 images.  The results, shown in Figure~\ref{fig:topbottom100}, are striking.  StyleGAN3 shows a remarkable bias for lighter faces and brighter images, with some apparent color bias: pink hair appears in many of the top-scoring images.  Since pink hair is uncommon in the dataset, the color bias \textit{cannot} be entirely explained by similar colors' frequency in the training data; this is, rather, pathological behavior more like ``hallucinations'' in other domains.  It may be due to a learned association with lighter skin tones, but despite this quirk, the lowest-scoring images are all dark. 

\subsection{Annotator Ratings of Top and Bottom 100 Faces}
Skin tone does not necessarily equate to ``race'' as a social category. To better measure the racial makeup of the faces at the tail ends of the discriminator score distribution, we have 300 raters rate a random sample of the top- and bottom-scoring of faces for perceived racial group. Results from the annotators in Figure~\ref{fig:racial categorization} indicate that images of Black people's faces were the plurality in the lowest 100 scoring faces, while images of white people's faces were the majority of the top 100 faces (and \text{none} were Black).

\subsection{Correlation with Image Luminance}

To better understand why this particular discriminator assigns faces with darker skin tones lower scores, we examine the relationship of luminance to the discriminator score.  In the GAN, images are encoded in an red-green-blue (RGB) color channel representation; from these, we can extrapolate a luminance value. The (photometric) \textbf{luminance} score~\cite{stone2016field} is a measure of light intensity for a particular area, the objective counterpart to subjective \textit{brightness}.  Relative luminance $Y \in [0,1]$ normalized luminance such that $1$ is reference white and $0$ is the darkest black.  Relative luminance can be calculated from the RGB channel representation of an image with Equation~\ref{eq:luminance} where $R$, $G$, and $B$ are the respective values for red, green, and blue, normalized to $[0,1]$~\cite{sugawara2014ultra}.
\begin{equation}
Y = (0.2126)R + (0.7152)G + (0.0722)B
\label{eq:luminance}
\end{equation}

\begin{figure}[]
\centering
\includegraphics[scale=0.5]{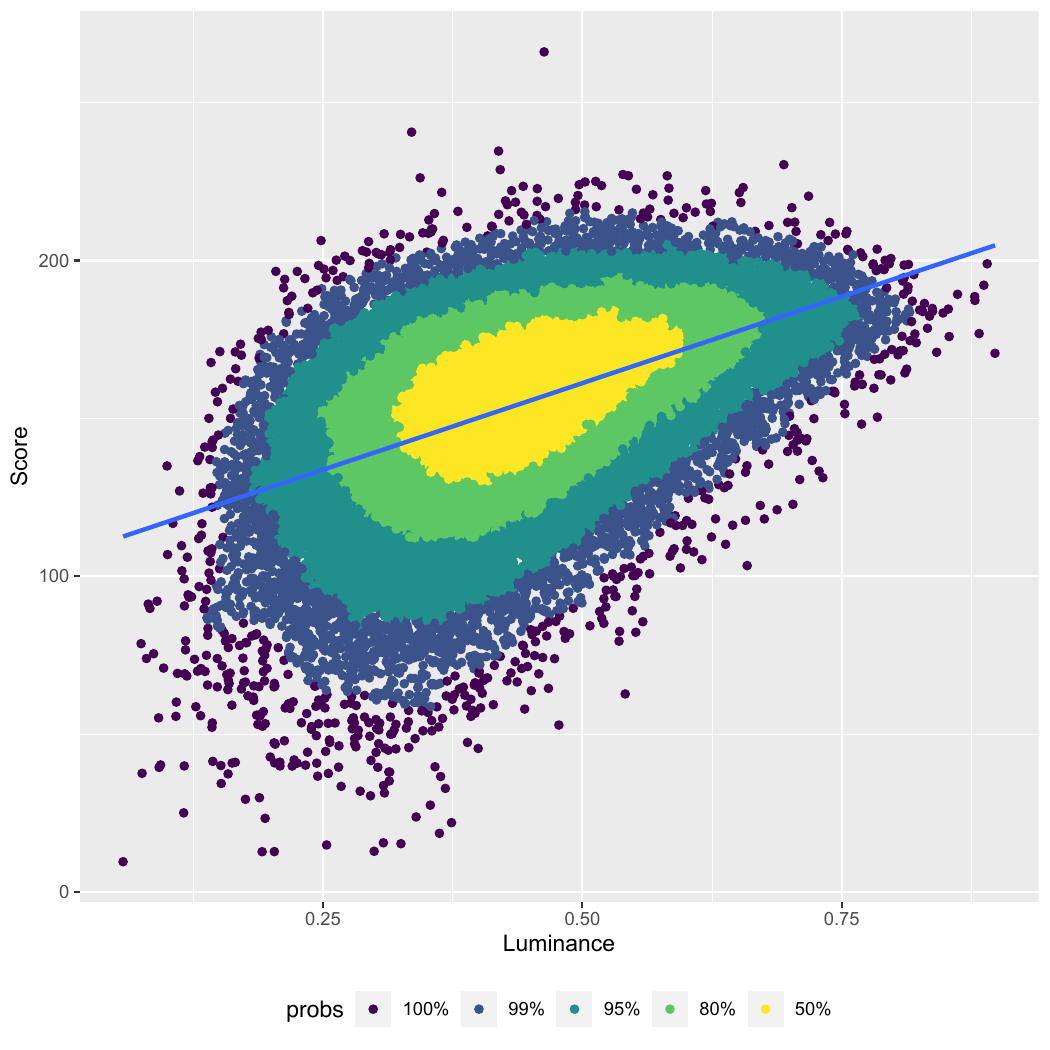}
\caption{Discriminator score as a function of luminance with a superimposed linear regression in the FFHQ dataset.  Colors show HDRs.  Scores increase  linearly with luminance, ut the 95\% HDR shows that this is not because high luminance images are especially common. This can be seen even more clearly in Figure~\ref{fig:luminance_dist}.}

\label{fig:luminance_correlation}
\end{figure}

We calculate the relative luminance of every image in  FFHQ and cross-reference this with the score assigned to the image by the discriminator. Figure~\ref{fig:luminance_correlation} shows a clear linear relationship between the luminance value and the score assigned to the image by the discriminator~(Figure~\ref{fig:luminance_correlation}).\footnote{A 95\% HDR or HDI indicates that 95\% of the data or distribution falls within the range, i.e., 95\% probability that a point falls within this region.}

\begin{figure}[t!]
\centering
\includegraphics[scale=0.4]{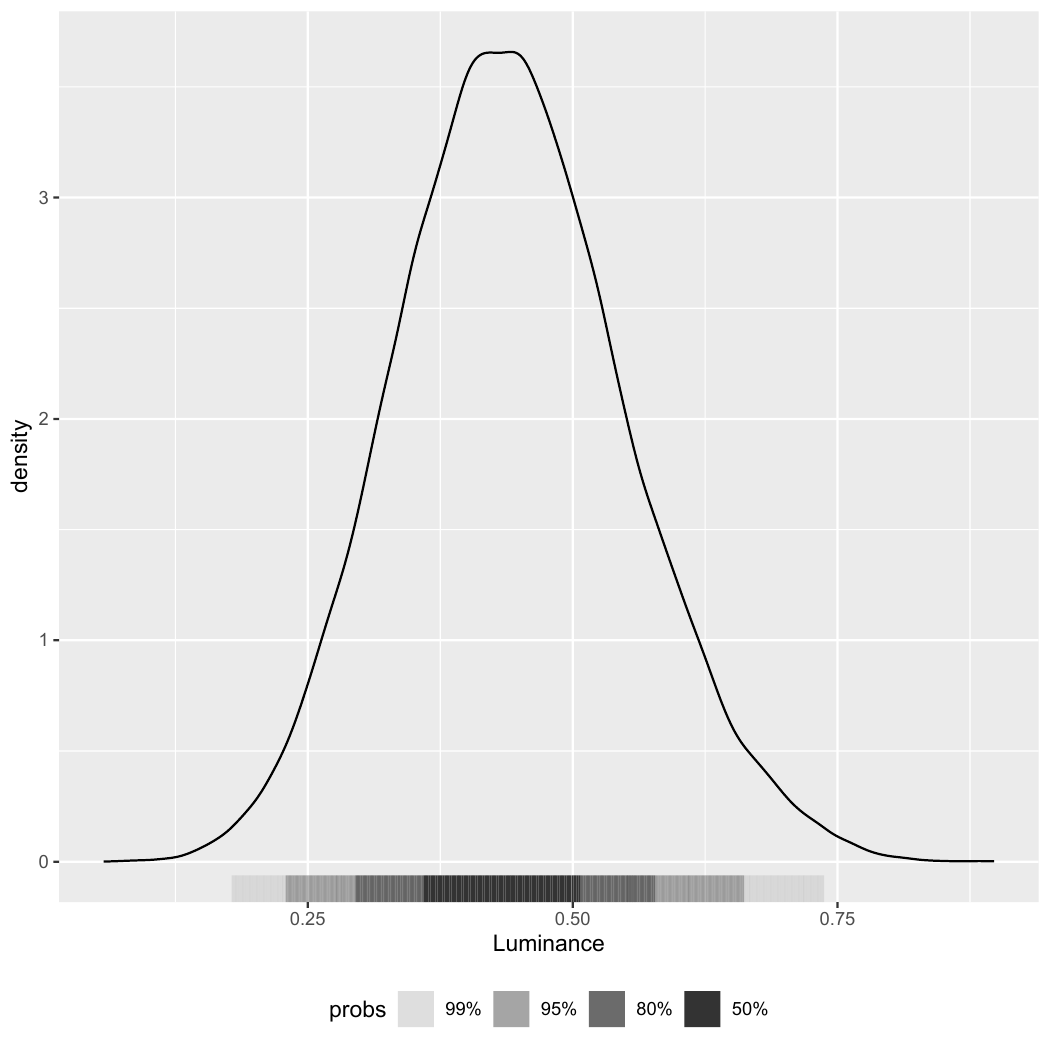}
\caption{Distribution of luminance over the training dataset.  The 95\% HDI shows that that vast majority of the data do not have especially high luminance, so this cannot explain the model's preference for it.}

\label{fig:luminance_dist}
\end{figure}

\subsection{Correlation with Skin Tone}

\begin{figure*}[t]
     \centering
     \begin{subfigure}[t]{0.45\textwidth}
         \centering
         \includegraphics[width=\textwidth]{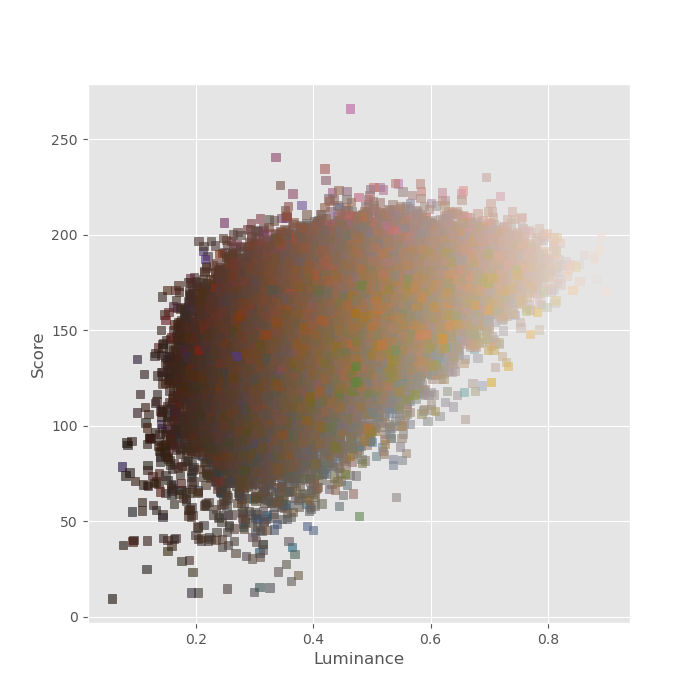}
         \caption{This is the same correlation shown in Figure~\ref{fig:luminance_correlation}, but with each of the 60,000 squares displaying the mean color of the image.  Along the $y$-axis, darker tones are generally scored lower at any given $x$ coordinate.  Cooler hues are penalized, and red is rewarded, with several red outliers. The variance in score clearly correlated with the variance in color.} 
         \label{fig:mean_color_scatter}
     \end{subfigure}
     \hfill
     \begin{subfigure}[t]{0.45\textwidth}
         \centering
         \includegraphics[width=\textwidth]{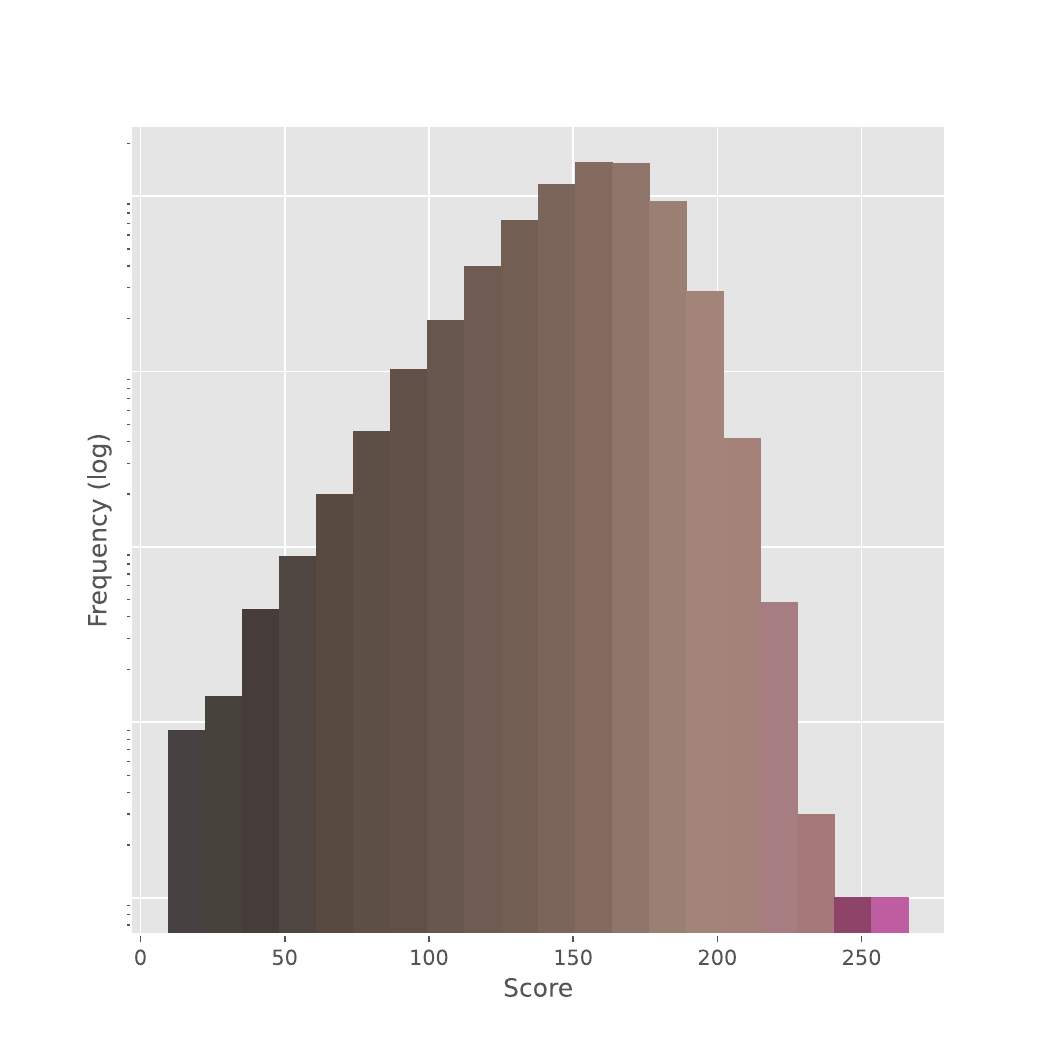}
         \caption{60,000 FFHQ photos binned by mean discriminator score. The color of each bar is the mean color of the images within the given score range. The smooth transition from darker to lighter skin colors is clear. Darker skin tones are dispreferred, but red is peculiarly rewarded. The right extremum, though a minuscule part of the training data and not a typical skin color, has exceptionally high scores. This appears pathological.}
         \label{fig:color_histogram}
        
     \end{subfigure}
     \caption{Color composition of the images is highly correlated with  score, beyond that which can be accounted for by luminance alone.  Though luminance is correlated with higher score, so is redness.  We cannot account for this with the composition of the training data, since this \textit{is} the training data, and many of the highest scoring colors occur \textit{less} frequently.}
     \label{fig:color_charts}
\end{figure*}

Luminance results indicate that the discriminator is assigning higher scores to photos with higher luminance. But higher luminance does not necessarily indicate with lighter skin tones.  To the extent that the camera is able to photograph darker skin tones we also sought to examine whether higher scores would be correlated with lighter skin tones, which correlates with (but is not fully explained by) luminance. Indeed, results indicate that lighter skin tones were associated with higher discriminator scores~(Figure~\ref{fig:color_charts}).

\subsection{Study 1 Discussion}
We see evidence that the discriminator scores faces with lighter skin tones higher (i.e., as more likely to be a face). Furthermore, although properties of the image such as luminance accounted for some of the variation in discriminator scores, Figures~\ref{fig:luminance_correlation}~and~\ref{fig:luminance_dist} indicate that luminance alone cannot explain the variation in scores. Although the findings from FFHQ indicate that Black faces are over-represented in the bottom 100 faces and white faces over-represented in the top 100, the racial makeup of the tails of this distribution depend on the variables of interest.

\section{Study 2: Scoring Novel, Annotated Faces}\label{sec:study2}

Study 1 (Section~\ref{sec:study1}) indicates that the discriminator is systematically biased against face images with darker skin in its training data. However, the racial categorization of these faces is the dependent variable in Study 1.  In Study 2, we now examine whether faces \textit{a priori} labeled as Black, white, or Asian are scored differently by the GAN.

\subsection{Labeled Faces Data Collection}

While the correlations with luminance (Figure~\ref{fig:luminance_correlation}) and color (Figures~\ref{fig:topbottom100}~and~\ref{fig:color_charts}) are compelling, we wish to know the degree to which the GAN's discriminator penalizes faces based on race and gender.

To examine this, we use Google Image Search to find faces of men and women of each of three perceived races: Asian, Black, and white. We recruit participants from the Prolific platform  to label both these images and the top and bottom 100 images from Study 1 for perceived race with choices determined by US Census categories. Of interest for our purposes are the categories \textit{Black or African American}, \textit{Asian or Pacific Islander}, and \textit{White}.  We ask participants for categorical labels for race.  We also elicit ratings for skin tone,\footnote{High numbers indicate darker tones.} hair length (\textit{long} or \textit{short}, as well numerical rating) as a cue for gender typicality~\cite{freeman2008will}, Asiocentricity, Afrocentricity, and Eurocentricity\footnote{See \cite{maddox2022cues} for a review of psychological research examining how facial phenotypes influence perception}. All scales had a 1-7 response scale; in general 1 = strong disagreement and 7 = strong agreement, though exact labels vary depending on the trait. We use the crowdsourced categorical labels for race to verify the authenticity of the labels and remove from the data those for which there is less than 2/3 agreement among raters.  We then manually inspect for obviously mislabeled images, which we remove. From those that remain, for each gender, we have at least 50 images for each race (50 with long hair, and 50 with short), leaving us with at least 300 images per labeled gender.\footnote{We have 1043 images total.}
\footnote{For the initial search, we use the following search terms, substituting the appropriate gender and race: \texttt{[race] [gender] with long hair}, \texttt{[race] [gender] with buzz cut}, and \texttt{Black men with locs}. We use this for Black men because \texttt{Black men with short hair} returned ambiguous and inconsistent results due to differing perceptions about what constitutes ``short hair'' for Black men.  These search terms were chosen so that we would have a variety of faces with different hair styles.} Since prior research has shown that the ``average face''generated by GANs is a white woman~\cite{salminen2020analyzing}, we hypothesized that more gender-atypical hair styles would attenuate the results.  We crop images so that they only contain the face.

\subsection{Results on Labeled Faces}
\label{sec:labeled_faces_results}
\begin{figure}[]

\centering
\includegraphics[width=0.5\textwidth]{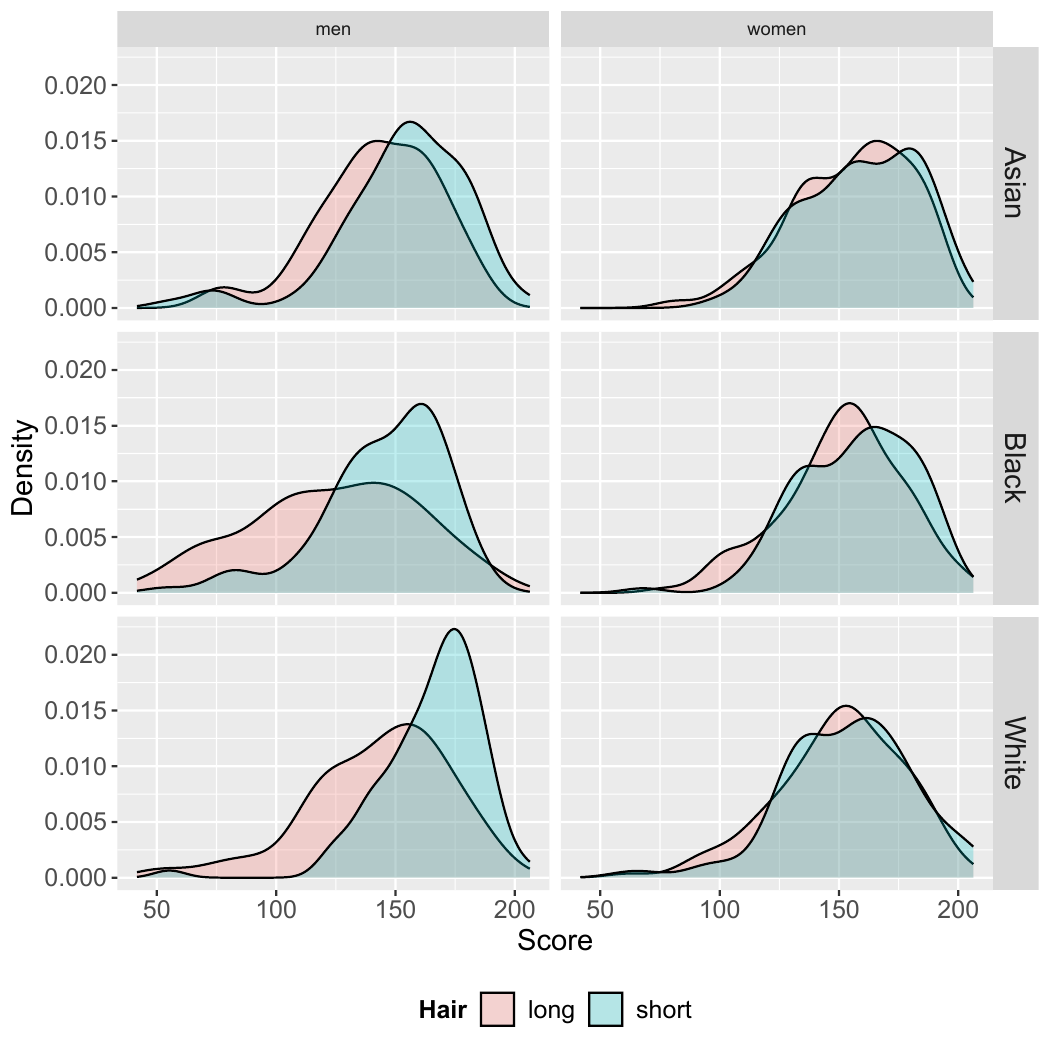}
\caption{Density plots for the faces labeled men  and women for each race. White men with short hair were scored substantially higher than those of any other group, though those with long hair are penalized heavily. Black men with long hair have the highest entropy distribution.  The differences in score for women's faces are mostly nonexistent. Note the consistent long left tails for women.}
\label{fig:hair_dists}
  
\end{figure}

A fair discriminator would not show substantial differences between each group.  However, this is not what we find. Figure~\ref{fig:hair_dists} shows clear differences between the discriminator-assigned scores for each race and hair style.  For men, long hair consistently penalizes the score: among men, effect is least pronounced for Asian men, but both black and white men see huge penalties, with long-haired black men's scores being the lowest and most entropic any group.
The faces of white men with short hair are heavily skewed toward the high end with a long tail on the low end.  Being non-white or having long hair both increase the entropy of the distribution for men.

Not so for faces labeled as women.  Asian and white women show no substantial difference in score based on hair type.  Black women, however, are penalized for having long hair.  Women's score distributions all have long tails on the left, which is not the case for men. It is surprising that the mode of white men's faces is so much higher and shifted to the right than that of women's faces, given that the ``average face'' generated by StyleGAN3 is a white women.

\section{Model-based Analysis}
\label{sec:modeling}
 \begin{figure}[]
   \centering
    \includegraphics[width=0.4\textwidth]{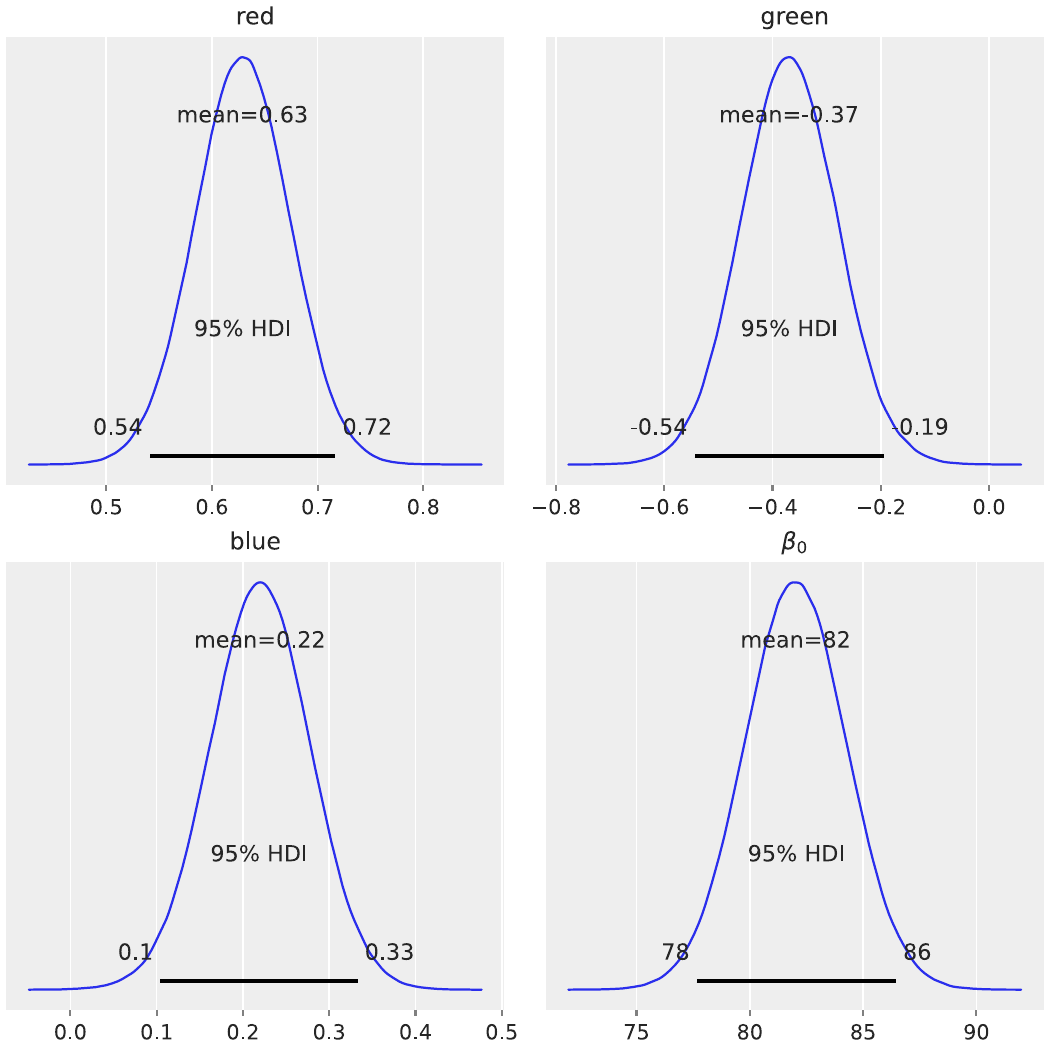}
    \caption{
    Coefficient posteriors for RGB values.
    }
    \label{fig:rgb_posteriors}
\end{figure}

 \begin{figure}[]
   \centering
    \includegraphics[width=0.45\textwidth]{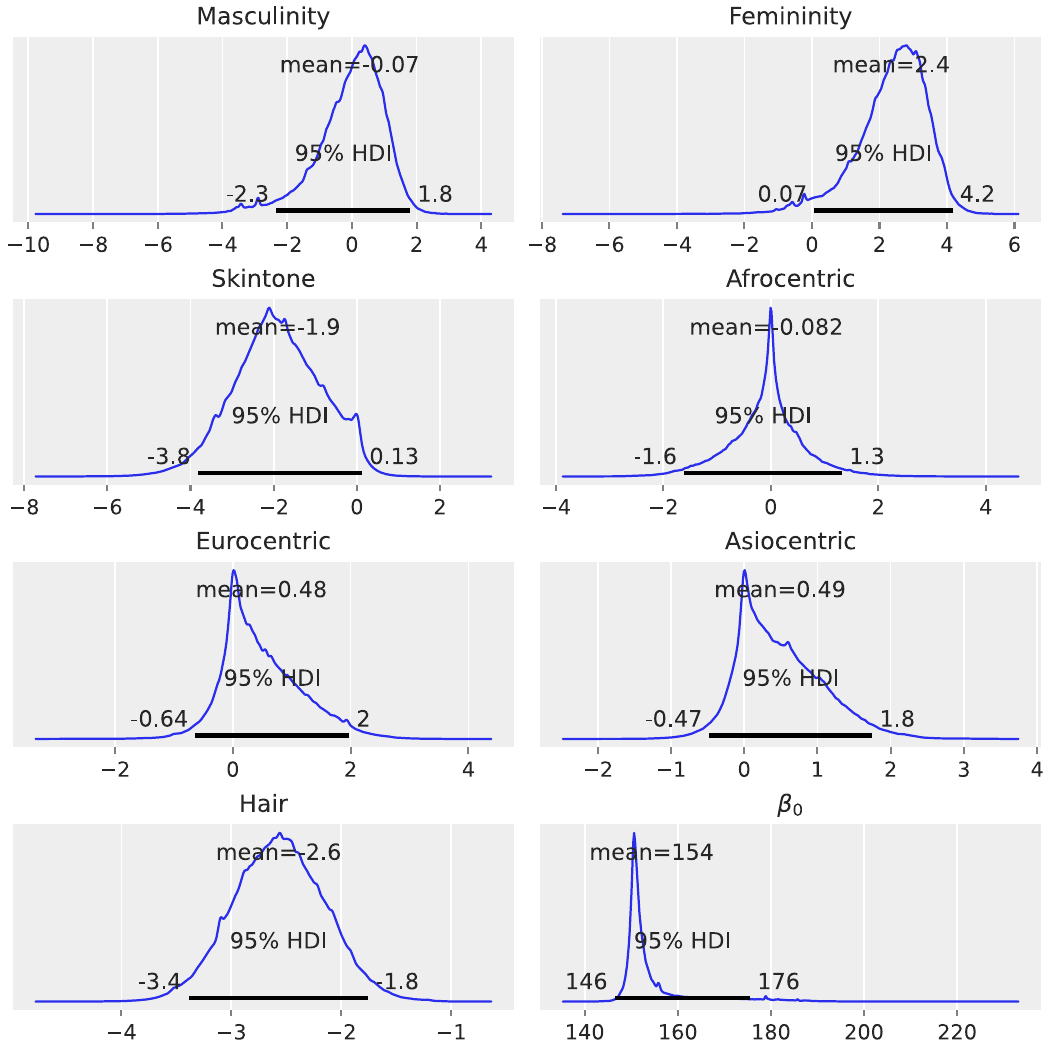}
    \caption{
    Weight posteriors for all elicited variables. Results conform to other experiments. Responses are all positive; so, relative ranking of similar categories is appropriate.
    }
    \label{fig:ordinal_posteriors}
\end{figure}

In this section, we examine the influence of color, skin tone, hair length, and the surveyed racial and gender characteristics of our dataset with Bayesian linear regression.  While we clearly cannot fully specify the salient variables of a CNN-based classifier with a linear model, we can examine general trends to interrogate the aggregate effects of such features, e.g., skin color, image content, and perceived race.  Since many of these variables are correlated (e.g., race and color), we view this as corroborating evidence.

\subsection{Bayesian Linear Regression on Colors}
We use an average of ten MCMC chains of 50,000 samples and 1,000 warm-up samples each (510,000 total) with the NUTS~\cite{hoffman2014no} algorithm to compute the posterior $p(\boldsymbol\beta|\mathbf{x}, y)$, where $\mathbf{x}$ is the data vector and $y$ is the discriminator score.
 The posteriors over weights indicate the relative contribution of each variable to the overall score according to the model, while the  shape of the curves and the HDI provide a sense of the range of credible values for the weights. 

\subsection{Preference for Color}
We begin by examining the contribution of image RGB means. We use a hierarchical model with separate hyperpriors on prior parameters. Let $C=\{r, b, g, 0\}$ be the set of all feature names (where 0 is the intercept feature).
For the discriminator score, let the likelihood
\begin{equation}
\begin{aligned}
    y &\sim \mathcal{N}(\mu_{\text{score}}, \sigma_{\text{score}}), \text{ where } \\
    \mu_{\text{score}} &= \beta_0 + \beta_r x_r + \beta_g x_g + \beta_b x_b,  \\
    \sigma_{\text{score}} &\sim \text{Half-}t(\sigma=1, \nu=10), \\
    \sigma_{c} &\sim \text{Half-}t(\sigma=1, \nu=10), \forall c\in C,\\
    \beta_0 &\sim \mathcal{N}(\mu=\bar{x}_{\text{score}},\sigma=\sigma_0), \\
    \beta_r &\sim \mathcal{N}(\mu=0, \sigma=\sigma_r), \\
    \beta_g &\sim \mathcal{N}(\mu=0, \sigma=\sigma_g), \\
    \beta_b &\sim \mathcal{N}(\mu=0, \sigma=\sigma_b). 
\end{aligned}
\end{equation}

 These fairly neutral, zero-centered weight priors leave us agnostic about the contribution of any particular color.\footnote{While we have presented evidence of bias for red in previous sections, for the purposes of this analysis, we feign ignorance.} We use the empirical score mean to center the intercept's prior.  Figure~\ref{fig:rgb_posteriors} shows a strong preference for red, a less pronounced preference for blue, and penalizes green.

\subsection{Bayesian Regression on User Ratings}
We also perform a robust Bayesian linear regression on the numeric responses we elicited.
 We use a $t$-distribution prior on the coefficients due to its heavy tails, robust to noise, and further place hyperpriors on a $\nu_\beta$ parameter shared among the weight priors to accommodate outliers. We also place hyperpriors on each standard deviation and mean.  We assume that masculinity and femininity share a variance and reasonably set $\nu=10$ in the standard deviation hyperprior.
 Let $C=\{\textit{Afrocentricity}, \textit{Asiocentricity}, \textit{Eurocentricity}, \textit{hair length},\\ \textit{masculinity}, \textit{femininity}\}$ be the set of features. Then,

\begin{equation}
    \begin{aligned}
    y &\sim \mathcal{N}(\mu_{\text{score}}, \sigma_{\text{score}}), \text{ where } \\
    \mu_{\text{score}} &= \beta_0 + \boldsymbol\beta^\top \mathbf{x},  \\
      \sigma_{\text{score}} &\sim \text{Half-Cauchy}(\gamma=10), \\  
     \sigma_{c} &\sim \text{Half-}t(\sigma=1, \nu=10),  \forall c\in C, \\
     \mu_c &\sim \mathcal{N}(\mu=0, \sigma=\sigma_c), \forall c\in C \\
    \nu_\beta &\sim \text{Half-Cauchy}(\gamma=2), \\
     \beta_{c} &\sim t(\nu=\nu_\beta, \mu=\mu_c, \sigma=\sigma_c), \forall c\in C.
\end{aligned}
\end{equation}

MCMC chains converge well and consistently under multiple runs. Figure~\ref{fig:ordinal_posteriors} shows the averaged posteriors over the $\beta$ weights, which conform to expectations given Studies 1 and 2. Since responses are all positive, relative ranking of comparable categories is informative, and negative values indicate strong dispreference in general. Masculinity is neutral (leaning negative), while femininity's 95\% HDI is positive and darker skin's HDI is almost all negative. Eurocentricity has a positive mean and a strongly positive HDI, while Afrocentricity's weight is extremely low by comparison; both distributions are quite Laplacian in the aggregate. Long hair and dark skin-tone are by far the most detrimental to one's score.
\subsection{Supplemental Analysis}

\begin{figure}[]
    \centering
    \includegraphics[width=0.5\textwidth]{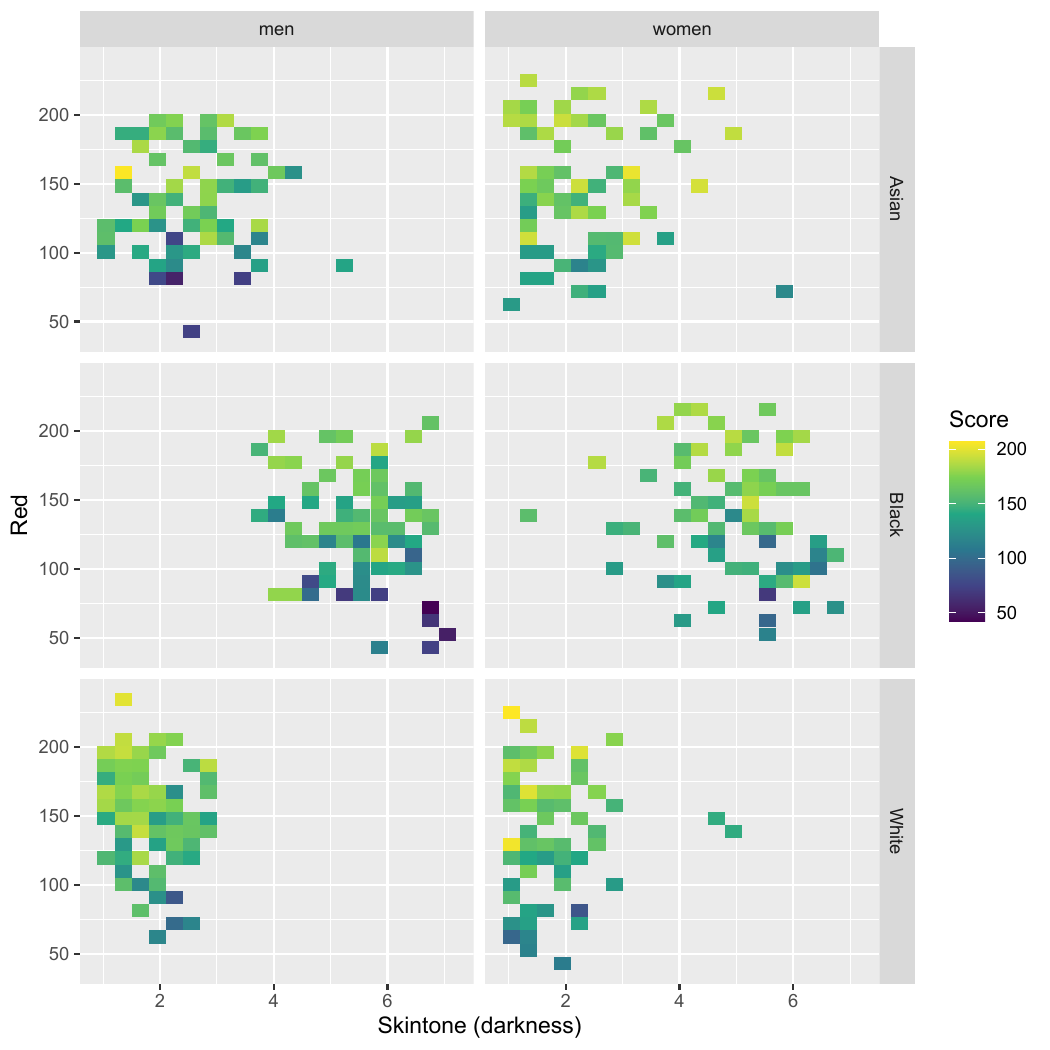}
    \caption{
    Binned interactions of skin-tone and redness. Black faces (and men) have more low scores. Redness does not fully explain the difference.
    }
    \label{fig:red_skin_grid}
\end{figure}

Our Bayesian regressions provide a bird's eye overview of the predictive qualities of the variables of interest and the credibility of our estimates.  There are, however, many possible interactions (e.g., Black men with long hair and light skin), and it is impossible to examine them all here. 
We can, however, examine the \textit{most} salient attributes by using a decision tree regression on a balanced dataset with exactly 50 labeled images for each race and gender, randomly sampled from the larger one. While there is some randomness, splitting nodes are chosen based on reduction in Gini impurity, which can be viewed as normalized information gain~\cite{biro2020gintropy}. We are not interested in training our own discriminator; instead, we can use a depth-limited decision tree to rank attributes by importance for minimizing mean squared error. The most discriminating features will appear higher up the tree. We train two depth-5 trees with all numerical features (RGB, ratings) and inspect the splitting nodes. Only a handful of features are used at this depth: \textit{red} (root, depths 1-4), \textit{blue} (2-4), \textit{green} (4), \textit{masculinity} (2-4), \textit{Eurocentricity} (3),  \textit{Afrocentricity} (4), and \textit{skintone} (4).
All nodes align with the trends shown in our other analyses but offer more evidence that not everything reduces to the effect of color means.  Further evidence for this can be seen in Figure~\ref{fig:red_skin_grid}, which suggests nonwhite men (and Black men in particular) may be penalized more than others.

\section{Conclusion and Future Work}

Across two separate face datasets, we find that the discriminator in the FFHQ-trained StyleGAN3 model is consistently biased in favor of white people. We see evidence of this in Study 1 (Section~\ref{sec:study1}), where the majority (74\%) of the highest-scored 100 images were white. Notably, although luminance and lighter skin tones account for some of the variation in discriminator scores, they could not fully explain the discriminator's bias. In Study 2 (Section~\ref{sec:study2}), we examined the explanatory power of stimulus race, again finding evidence that the discriminator favored white faces. Looking at the result of the Bayesian regression analyses, we find that rated facial Eurocentricity is a strong positive predictors of discriminator score.

We also found evidence that hair length mattered. In Study 2, images with longer hair were generally scored lower by the discriminator, though this was particularly prominent among images of Black people and men (and Black men specifically). This is interesting in the context of white women being the most commonly generated group by GANs~\cite{salminen2020analyzing}. Given this context, we hypothesized that short hair might be penalized relative to long hair but found the opposite. One reason for this might be that the discriminator is particularly focused on contours of the face, which long hair may be more likely to obscure. Future work can examine whether short hair styles that obscure features of the face (e.g., bowl haircuts) might also be scored lower by the discriminator.

These studies provide some insight into the biases and pathologies of GANs and also suggest some future directions. The most obvious next step is to examine to what extent these results generalize to variations in model type, initial conditions, and training data.  Additionally, if the GAN is particularly sensitive to image luminance, then perhaps there are key features must be able to detect to decide that the image contains a face. Infants do something similar, using perceptual cues to engage in face-processing~\cite{turati2005three}. Yet as 3-months-of-age, infants shift away from pure perceptual processing and shift to engaging in more configural processing of faces~\cite{turati2005three}.  Incorporating some of these principles of face-processing through metacognition may not only enhance GAN face generation but also minimize demographic biases. Similar paradigms using metacognition have been employed for object recognition~\cite{berke2021learning}. 

GANs have huge promise in generating photorealistic images of faces for use in a variety of applications. Failure to account for its demographic biases, however, will only further perpetuate existing social inequalities. 
\section{Acknowledgements}
This work is funded by \href{https://www.nsf.gov/awardsearch/showAward?AWD_ID=2210142}{NSF Award 2210142}, EAGER: DCL: SaTC: Enabling Interdisciplinary Collaboration: Evaluating Bias In The Creation and Perception of GAN-Generated Faces.

\bibliographystyle{abbrv}
\bibliography{arxiv_draft}  






\end{document}